\pdfoutput=1

\documentclass[11pt,final]{article}

\usepackage[final]{acl}

\usepackage{times}
\usepackage{latexsym}

\usepackage{graphicx}
\usepackage{booktabs}
\usepackage{multirow} 
\usepackage{amssymb}

\usepackage[T1]{fontenc}

\usepackage[utf8]{inputenc}

\usepackage{microtype}

\usepackage{inconsolata}

\usepackage{graphicx}

\title{Explicit vs. Implicit Biographies: Evaluating and Adapting LLM Information Extraction on Wikidata-Derived Texts}

\author{
 \textbf{Alessandra Stramiglio\textsuperscript{1,2}},
 \textbf{Andrea Schimmenti\textsuperscript{1}}, \\
 \textbf{Valentina Pasqual\textsuperscript{3}}, 
 \textbf{Marieke van Erp \textsuperscript{4}},
 \textbf{Francesco Sovrano \textsuperscript{5}}
 \textbf{Fabio Vitali\textsuperscript{1,3}}
\\
\\
 \textsuperscript{1} DISI Departement of Computer Science and Engineering, University of Bologna, Italy, \\ 
 \textsuperscript{2} Automobili Lamborghini SpA, Sant'Agata Bolognese, Italy, \\
 \textsuperscript{3} Digital Humanities Advanced Research Center (/DH.arc),  \\
Department of Classical Philology and Italian Studies, University of Bologna, Italy \\
 \textsuperscript{4} KNAW Humanities Cluster, DHLab, Amsterdam, the Netherlands
\\
 \textsuperscript{5} ETH Zurich, Collegium Helveticum, Switzerland
\\
 \small{
    \textbf{Correspondence:} \href{mailto:a.stramiglio@unibo.it, andrea.schimmenti2@unibo.it, valentina.pasqual2@unibo.it, fabio.vitali@unibo.it, marieke.van.erp@dh.huc.knaw.nl}{\{a.stramiglio, andrea.schimmenti2, valentina.pasqual2\}@unibo.it, marieke.van.erp@dh.huc.knaw.nl}
 }
}

\begin{document}
\maketitle

\begin{abstract}
Text Implicitness has always been challenging in Natural Language Processing (NLP), with traditional methods relying on explicit statements to identify entities and their relationships. From the sentence "Zuhdi attends church every Sunday", the relationship between Zuhdi and Christianity is evident for a human reader, but it presents a challenge when it must be inferred automatically.
Large language models (LLMs) have proven effective in NLP downstream tasks such as text comprehension and information extraction (IE). 

This study examines how textual implicitness affects IE tasks in pre-trained LLMs: LLaMA 2.3, DeepSeekV1, and Phi1.5. 
We generate two synthetic datasets of 10k implicit and explicit verbalization of biographic information to measure the impact on LLM performance and analyze whether fine-tuning implicit data improves their ability to generalize in implicit reasoning tasks.  

This research presents an experiment on the internal reasoning processes of LLMs in IE, particularly in dealing with implicit and explicit contexts. The results demonstrate that fine-tuning LLM models with LoRA (low-rank adaptation) improves their performance in extracting information from implicit texts, contributing to better model interpretability and reliability.
The implementation of our study can be found at \href{ https://github.com/aschimmenti/xAi-KE-ImplicitKnowledge}{ImplicitKnowledge}

\end{abstract}

\section{Introduction} \label{sec:introduction}
Information Extraction (IE) seeks to identify, classify, and represent entities from unstructured textual sources. Large Language Models (LLMs) significantly improved Natural Language Processing (NLP) performance in IE, demonstrating remarkable capabilities in tasks such as text comprehension, classification, Named Entity Recognition (NER) and Relationship Extraction (RE) \cite{niklaus-etal-2018-survey,10152821-openie-zeroshot}. Conventional approaches (e.g., rule-based, deep learning) predominantly rely on explicit statements to extract entities, relations, and events \cite{alt-etal-2020-tacred}. However, real-world texts also convey information implicitly, requiring inferential processing to derive the intended meaning. 

Implicit meaning arises when information is conveyed indirectly through linguistic and cognitive mechanisms rather than explicitly stated where contextual reasoning and pragmatic inference are required to perform correct interpretations \cite{yule1996pragmatics,VyvyanCognitive2012,Fischer_2017}. The sentence "Zuhdi attends \textit{church every Sunday}" suggests Zuhdi is likely Christian, as this inference is drawn from a religious frame, requiring additional knowledge to make sense of what is not explicit. Similarly, the statement "Sarah received her degree from Oxford University on \textit{June 15, 2010}, and celebrated her \textit{20th birthday the same day}" implies she was born on June 15, 1990, establishing a temporal entailment that is not explicitly stated.

Biographical texts present a challenging case for information extraction due to their reliance on the implicit use of language. Although these texts do not require specialized domain knowledge for comprehension, they present a moderate level of complexity \cite{tint2024expressivityarenallmsexpressinformation}. Such complexity arises from the relationships between entities, temporal dependencies, and occupational references, which are often inferred through contextual cues rather than explicitly stated.

This study investigates the impact of textual implicitness on LLM-based IE tasks, tackling two main research questions (RQs): 
\begin{itemize}
    \item RQ1: How do implicit and explicit verbalizations affect LLM performance in information extraction tasks?
    \item RQ2: {How does exposure to implicit data during fine-tuning affect an LLM’s ability to generalize to implicit reasoning tasks?}
\end{itemize}

Since LLMs often exhibit difficulty in extracting information from implicit contexts \cite{tint2024expressivityarenallmsexpressinformation}, we explore whether fine-tuning can mitigate this difficulty. Specifically, we investigate the impact of fine-tuning on well-known models from the community such as LLama3.2 \cite{llama3.2}, DeepSeekV1 \cite{deepseekai2025}, and Phi1-5 \cite{Phi15}. This is particularly relevant for scenarios where critical information is conveyed implicitly rather than explicitly. Our findings contribute to improving model reliability and expanding potential applications.

We focus on two datasets, one explicit and one implicit, containing natural language descriptions of people's biographies. The texts were synthetically generated starting from a Wikidata triple dataset. By fine-tuning models on implicit patterns which mimic real-world scenarios, we assess their ability to extract information from implicit texts.

This contribution is summarized as follows: Section \ref{sec:background} lays the background of this work, Section \ref{sec:method} presents the adopted methodology to provide an answer to RQ1 and RQ2. Our results are presented and discussed in Section \ref{sec:discussion}. Finally, \ref{sec:conclusion} outlines our final remarks and future works.

\section{Background and Related Work} \label{sec:background}
IE focuses on structuring data, e.g. in the form of a \textit{triple}, where two arguments are connected through a relation. Usually, this takes the form of a triple composed of a \textit{subject}, a \textit{predicate} and an \textit{object}, as <s, p, o>  \cite{niklaus-etal-2018-survey}. This task is usually defined as Relationship Extraction (RE).
One inner distinction is the difference between Closed and Open RE. Closed RE focuses on finding arguments given one or more constraints (e.g., <s, p, ?o> - where the object is the only unknown value), while Open RE looks for any potential triple in a text (e.g. <?s, ?p, ?o>) instead.
Traditional RE models primarily identify triples where elements (subjects, predicates, and objects) have explicit textual mentions. These models are trained to recognize explicit linguistic markers (such as verbs functioning as predicates) but often struggle with implicit relationships that require common sense knowledge or deeper natural language understanding \cite{pei-etal-2023-abstractive}. Pre-trained Language Models (PTLMs) and LLMs represent the state-of-the-art for unsupervised Open IE tasks \cite{10152821-openie-zeroshot}, as they can process implicit information more effectively than previous approaches.

The role of implicit and explicit knowledge has been extensively studied in cognitive science. According to Dienes and Perner's theory, implicitness arises when information is conveyed indirectly through the functional use or conceptual structure of explicit representations, rather than being directly represented \cite{Dienes_Perner_1999}. 

In RE, being able to identify relationships with different levels of explicitness presents a significant challenge. 
LLMs have shown that while these models can effectively process explicit information, they still struggle with implicit knowledge that requires commonsense reasoning \cite{Ilievski2024HumanCentricAW}. 
The dimensions of implicit relationships can vary significantly based on:
\begin{itemize}
    \item The level of inference required (from simple logical deduction to complex contextual reasoning)
\item The type of background knowledge needed (from common facts to domain expertise)
\item The cultural and temporal context necessary for understanding
\end{itemize}

The degree of implicitness in information also directly impacts the certainty with which models can retrieve and reason about that information. While explicit statements can be processed with high confidence, implicit information introduces varying levels of uncertainty that models must learn to handle appropriately. 
Datasets for RE usually prioritize explicitly stated information. For instance, RED \cite{huguet-cabot-et-al-2023-redfm-dataset}, a widely used RE dataset, focuses on extracting triples that directly match sentences found in the text. RED provides entity types and relationships without enforcing additional structural constraints, such as predefined categories for entities or specific restrictions on how relationships should be formed - the domain and range of the predicate (see Table \ref{tab:red_example})\footnote{The dataset entry was extracted from \url{https://huggingface.co/datasets/Babelscape/REDFM}}.

\begin{table}[ht!]
    \centering
    \begin{tabular}{p{2cm}p{2cm}p{2cm}}
        \toprule
        \textbf{Subject} & \textbf{Predicate} & \textbf{Object} \\
        \midrule
        Émilie Andéol & sport & judo \\
        \midrule
        \multicolumn{3}{p{6cm}}{\textit{Source text:} "Émilie Andéol [...]is a French judoka competing in the women's +78 kg division."} \\
        \bottomrule
    \end{tabular}
    \caption{Example of a RED dataset triple-sentence pair}
    \label{tab:red_example}
\end{table}

This uncertainty increases proportionally with the degree of inference required to extract the information. 
For example, consider these statements about Gaia, from which we want to draw a statement about her occupation:
\begin{itemize}
\item "Gaia works as a doctor at City Hospital"
\item "Gaia wears a white coat and sees patients daily" 
\item "Gaia ran through the emergency room corridor, quickly reviewing charts"
\end{itemize}
All the statements convey the same information with a different degree of implicitness. The first information is explicit, as the sentence is shaped similarly to the <s, p, o> structure, where ?o is equal to the attribute of the verb \textit{works}. The occupation, in the second case, is described by the daily routine of the occupation itself (metonymy). In the third case, the information is hid completely: even a human would not be able to discern her occupation with certainty. Different people could rush through an emergency room with a chart in hand, not necessarily a doctor (unless additional context provides more clues). The less a statement is explicit, the more uncertainty builds up.

LLMs seem to struggle with processing implicit information, \cite{becker-etal-2021-reconstructing}  we sought to better understand whether this limitation arises from the model's architecture or its training data. Specifically, we investigate whether this issue reflects \textit{aleatoric uncertainty}—stemming from inherent unpredictability in language—or \textit{epistemic uncertainty}, where performance is limited by the model's exposure to certain distributions during training \cite{hullermeier2021aleatoric}.

For this purpose, we fine-tune LLMs on RE tasks that include varying degrees of implicitness. This approach allows us to probe the model's generalization capacity and assess whether performance improvements emerge from better learning of the input-output mappings or from increased familiarity with implicit patterns. Rather than focusing on post-hoc interpretability methods \cite{XAI, molnar2022interpretable}, we position our analysis as a behavioral study of LLMs in the context of information extraction, with a focus on how implicitness affects extraction reliability.

In recent years, the development of robust IE systems has increasingly depended on the availability of high-quality data. However, for many domains, available datasets are limited in size, making data augmentation and synthetic dataset generation techniques gain traction as practical solutions. In NLP, for instance, methods such as back translation and synonym replacement have long been employed to expand parallel corpora \cite{augment_data_approaches}. More recently, Synthetic Dataset Generation has emerged as a strategy that leverages LLMs to create training data for smaller models, especially for tasks or domains with limited human-labeled examples \cite{gpt_synthetic_empirical}. This strategy has proven valuable in medical and low-resource contexts where annotation is both expensive and time-consuming \cite{chebolu-etal-2023-review}.

This synthetic data generation approach is particularly relevant for addressing the challenges of implicit RE discussed earlier. By generating diverse examples with varying degrees of implicitness, we can potentially improve model performance on the full spectrum of RE tasks—from explicit statements to those requiring complex inference and entailment. Typically, synthetic data is generated starting from a single prompt or a minimal set of guiding rules \cite{long-etal-2024-llms}, aiming to steer the model toward desired outputs. However, generating high-quality synthetic data for implicit relationships remains challenging, as it requires the generative model to simulate the complex reasoning processes that humans use to infer unstated connections.

\section{Method} \label{sec:method}

\begin{figure*}[h!]
    \centering
    \includegraphics[width=0.9\textwidth]{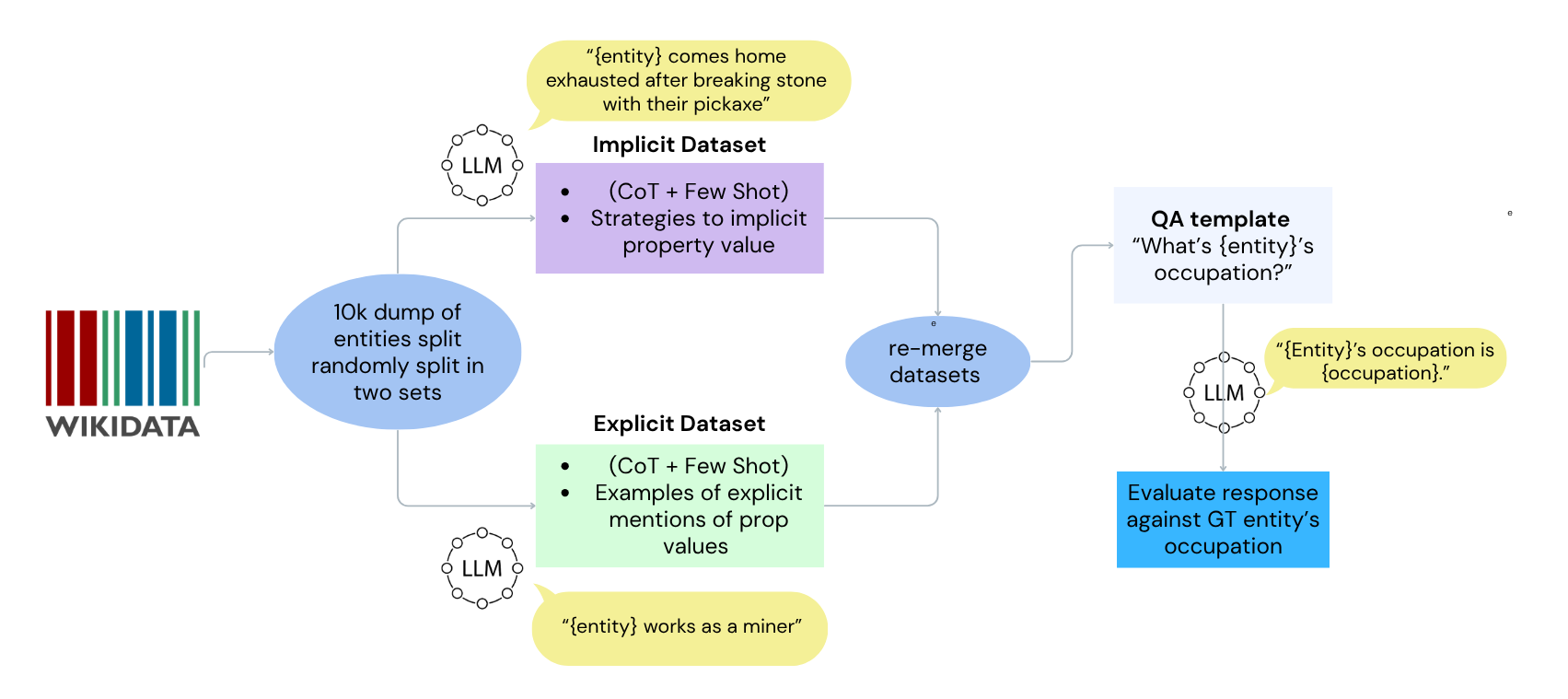}
    \caption{Dataset generation and Experiment setup}
    \label{fig:dataset_experiment_setup}
\end{figure*}

This section outlines the design of a controlled experiment conducted to examine the impact of implicit and explicit IE in the performance of an LLM, thereby addressing research questions RQ1 and RQ2. Figure \ref{fig:dataset_experiment_setup} summarizes the overall method employed to conduct this study.

First, a set of 10,000 random entities from Wikidata was extracted, specifically targeting entities of the Human class\footnote{The Wikidata class "Human" is identified via the  ID \href{https://www.wikidata.org/wiki/Q5}{Q5}}, e.g. \href{https://www.wikidata.org/wiki/Q21931962}{Vincent Rodriguez III}). The entities' biographical information\footnote{Represented in Wikidata as Statements (\url{https://www.wikidata.org/wiki/Help:Statements})} 
have been extracted via the Wikidata API, filtering out irrelevant information, such as identification parameters, visual references, and associated technical metadata. As shown in Table \ref{tab:example_descriptive_triples}, 14 triples describe relevant information about the biography of Vincent Rodriguez III (e.g., occupation, country of citizenship, sexual orientation), with 18 values. 
Our aim is to create two parallel sentences for each person, one that describes a fact or info about them explicitly, and the other implicitly. First, a random property is selected for each person. An example is shown in Table \ref{tab:example_descriptive_triples}, where the selected information for Vincent Rodriguez III is his occupation as a television actor.

\begin{table*}[h!]
    \centering
    \begin{tabular}{|p{3cm}|p{3cm}|p{3.7cm}|c|}
        \hline
        \textbf{Subject} & \textbf{Predicate} & \textbf{Object} & \textbf{Hidden info} \\
        \hline
        \multirow{18}{*}{\parbox{3cm}{Vincent Rodriguez \\ III}}
        & instance of & human & $\times$  \\
        \cline{2-4}
        & place of birth & San Francisco & $\times$  \\
        \cline{2-4}
        & sex or gender & male & $\times$  \\
        \cline{2-4}
        & given name & Vincent & $\times$  \\
        \cline{2-4}
        & \multirow{2}{*}{occupation} & actor & $\times$  \\
        \cline{3-4}
        & & \textbf{television actor} & \textbf{$\checkmark$}  \\
        \cline{2-4}
        & country of citizenship & United States & $\times$  \\
        \cline{2-4}
        & sexual orientation & homosexuality & $\times$  \\
        \cline{2-4}
        & date of birth & +1982-08-10T00:00:00Z & $\times$  \\
        \cline{2-4}
        & \multirow{2}{*}{educated at} & Pacific Conservatory of the Performing Arts & $\times$  \\
        \cline{3-4}
        & & Westmoor High School & $\times$  \\
        \cline{2-4}
        & family name & Rodriguez & $\times$  \\
        \cline{2-4}
        & \multirow{3}{*}{residence} & Daly City & $\times$  \\
        \cline{3-4}
        & & New York City & $\times$  \\
        \cline{3-4}
        & & North Hollywood & $\times$  \\
        \cline{2-4}
        & languages spoken, written or signed & English & $\times$  \\
        \cline{2-4}
        & native language & English & $\times$  \\
        \cline{2-4}
        & writing language & English & $\times$  \\
        \hline
    \end{tabular}
    \caption{Selected information about Vincent Rodriguez III. The Table comprehends all triples (subject, predicate, object) available for the entity Rogriguez III, excluding non-semantic information (e.g., resource identifiers, image links)}
    \label{tab:example_descriptive_triples}
\end{table*}
Then, all the information about that person alongside which information needs to become implicit becomes the input of the prompt. GPT-4o is instructed to generate two different sentences: an explicit one (similar to Wikipedia's straightforward style), and an implicit one where the same information is conveyed through narrative context and indirect references. 
The prompt uses Few-Shot learning (with 10 examples) and Chain-of-Thought. The examples for generating the implicit sentence are sentences with a paired rhetorical strategy (e.g. periphrasis, metonymy, deduction)\footnote{Refer to the following script for the complete prompt: \textit{prompt\_generation\_implicit.py}}.
Table \ref{tab:example_descriptions} presents the generated sentences about Vincent Rodriguez III. The selected property is stated explicitly in the first description, i.e. "he is a famous \textbf{television actor}", while in the second it is implied through a periphrasis (i.e. "\textbf{showcasing his talent in various television productions}").

\begin{table}[h]
    \centering
    \begin{tabular}{p{7cm}}
        \hline
        \textbf{Explicit Description} \\ \hline
        Vincent Rodriguez III, born on August 10, 1982, in San Francisco, has captivated audiences with his performances since his early days at the Pacific Conservatory of the Performing Arts. Residing in vibrant cities like New York and North Hollywood, he has embraced the world of entertainment; he is a famous \textbf{television actor}. \\ \hline
        \textbf{Implicit Description} \\ \hline
        Vincent Rodriguez III, born on August 10, 1982, in San Francisco, has captivated audiences with his performances since his early days at the Pacific Conservatory of the Performing Arts. Residing in vibrant cities like New York and North Hollywood, he has embraced the world of entertainment, \textbf{showcasing his talent in various television productions} that highlight his dynamic range and charisma. \\ \hline
    \end{tabular}
    \caption{Implicit and Explicit Descriptions about Vincent Rodriguez III}
    \label{tab:example_descriptions}
\end{table}


Finally, two IE tests were performed as Question-Answering. We test the model's ability to retrieve the implicit and explicit information (e.g. "What's Vincent Rodriguez III's occupation?"). As an additional rule, both answer can be considered valid, but "television actor" is counted as the better answer, being the more fine-grained answer compared to simply "actor". Indeed, IE on implicit sentences can exhibit reduced precision by retrieving only the hypernym "actor" rather its more specific hyponym, despite the presence of the modifier "television" as shown in Table \ref{tab:example_qa}.

\begin{table}[ht]
    \centering
    \begin{tabular}{p{3cm}p{1.5cm}p{1.5cm}}
        \hline
        \textbf{Question} & 
        \textbf{Explicit Answer} & \textbf{Implicit Answer} \\
        \hline
        What does Vincent do for a living? & Television actor & Actor \\
        \hline
    \end{tabular}
    \caption{Comparison of Explicit and Implicit Answers}
    \label{tab:example_qa}
\end{table}

\subsection{RQ1: Preliminary results and evalutation}\label{sec:preliminary_results}
Evaluation has been performed over answers provided on explicit and implicit descriptions. As postprocessing, we performed lemmatization to align model answers to the Wikidata vocabulary. Then, the semantic distance between the expected answer (e.g.,  \textit{Television actor}) and LLM-generated answers (e.g., \textit{Television actor} from explicit description and \textit{actor} from implicit description) has been computed trough BLEURT \cite{bleurt}.
We performed the Wilcoxon signed-rank test to assess whether the difference between the two distributions was statistically significant. This non-parametric test compares two related samples to determine if their population mean ranks differ. Applying the Wilcoxon signed-rank test, the two distributions are statistically significant considering a $pvalue < 0.05$.
Moreover, the percentage of NaN values given by the model when exposed to the implicit text is way higher, with a value of 14.60\% against 1.30\% of the explicit. These evaluations give us grounds to use the generated dataset in the evaluation of RQ2. Details of the test and the results can be found in the Github repository \href{ https://github.com/aschimmenti/xAi-KE-ImplicitKnowledge}{ImplicitKnowledge}.


\subsection{RQ2: Approach to Fine Tuning}
The evaluation of our preliminary results (Section \ref{sec:preliminary_results}) addresses RQ1, showing that statistically, the model struggles more with information extraction when sentences follow a pattern of implicitness. Hence our second research question (RQ2): \textit{How does exposure to implicit data during fine-tuning affect an LLM’s ability to generalize to implicit reasoning tasks?}
To demonstrate this from the dataset validated above, we decided to take a subset of it where we select only a few occupations. To choose them, we took the 5 most common occupations {‘actor’, ‘film actor’, ‘television actor’, ‘stage actor’, ‘film director’} in the property values, i.e. in the ground truth and the respective Implicit and Explicit sentences as shown in Table \ref{tab:example_descriptions}. 

\subsubsection{Experiment} \label{experiment}
The experiment explores whether fine-tuning an LLM model can improve its ability to perform IE on implicit instances by training it in different settings:
\begin{itemize}
    \item \textbf{Training on explicit IE, testing on explicit IE}. We expect it to work correctly, as it should be the easiest setting;
    \item \textbf{Training on implicit IE, testing on implicit IE.} Again, we expect it to perform well by training it directly on this task;
    \item \textbf{Training on both explicit and implicit IE, testing on both}, \textbf{one} for \textbf{explicit} and \textbf{one} for \textbf{implicit}. If trained together, is the model able to classify the two different sets correctly?
    \item \textbf{Training on explicit IE, testing on implicit IE}. From what we've seen above in RQ1, we expect this one to be the hardest task for the model as it requires to generalize the most.
\end{itemize}

\subsubsection{Models}
For the classification tasks in our study, we selected three models that are significant and widely recognized within the community. We selected LLaMA, DeepSeek, and Phi for our experiments based on their widespread adoption and their performance within NLP research. At the time of writing (\textit{April 2025}), both LLaMA and DeepSeek have shown substantial popularity, with 2.1 million and 1.8 million downloads respectively in last month on the \href{https://huggingface.co/}{Hugging Face} Platform, indicating broad usage and interest.

Although Phi models (developed by Microsoft) have comparatively fewer downloads ($\sim$ 100K), they remain a valuable inclusion due to their strong performance relative to their size. As highlighted by the Hugging Face model card \href{https://huggingface.co/microsoft/phi-1_5}{Hugging Face} and supporting benchmarks, Phi-1.5 achieves near state-of-the-art results among models with fewer than 10 billion parameters, making it a compelling lightweight alternative for evaluating instruction-tuned models.

Overall, our selection balances community adoption, model diversity and openness, and parameter efficiency, allowing for a robust and representative evaluation across the current LLM landscape.

\begin{itemize}
    \item \textit{meta-llama/Llama-3.2-1B:} Developed by Meta AI, this model is part of the Llama 3.2 collection of multilingual LLMs. It is optimized for multilingual dialogue use cases, including agentic retrieval and summarization tasks. \cite{llama3.2}
    \item \textit{DeepSeek-R1-Distill-Qwen-1.5B}: Developed by Deepseek AI, this is a distilled version of the DeepSeek R1 model. Given its cost-effectiveness and performance, it is a competitive choice for NLP tasks. \cite{deepseekai2025}
    \item \textit{microsoft/phi-1\_5} A transformer-based model from Microsoft, trained using the same data sources as Phi-1, augmented with new data. It shows similar state-of-the-art performance among models with less than 10 billion parameters.  \cite{Phi15}
\end{itemize}
Each of these models contains between 1 and 1.5 billion parameters, and they are hosted on the Hugging Face platform \cite{wolf2020huggingfacestransformers}. Given our necessity to test fine-tuned performance, we have chosen only open-source models, as we need access to the weights and structure of the models. This approach also ensures reproducibility.

\subsubsection{LoRA fine-tuning}
To build the classification model, we used finetuned LLMs with Low-Rank Adaptation (LoRA) \cite{hu2021loralowrankadaptationlarge} for the sequence classification task. LoRA \cite{hu2021loralowrankadaptationlarge} is a parameter-efficient fine-tuning technique that draws inspiration from studies on the intrinsic dimensionality of hyperparametrised models. Research by \cite{li2018measuringintrinsicdimensionobjective} and \cite{aghajanyan2020intrinsicdimensionalityexplainseffectiveness}  has shown that such models operate in a low intrinsic dimension, suggesting that vast parameter spaces can be efficiently navigated in a more compact subspace. Building on this insight, LoRA hypothesises that the weight changes required during model fitting also have a low ‘intrinsic rank’. Consequently, instead of updating all model parameters during fitting, LoRA introduces low-rank trainable matrices that approximate these weight changes. The overview of the parameters used is in Table \ref{tab:models_parameters} while the details are provided in Table \ref{tab:appendix_hyperparameters} in Appendix \ref{sec:appendix}

\subsection{Training Details}
We trained, in total, 9 classifiers, with different training for each of the three models, Llama-3.2-1B, DeepSeek-R1-Distill-Qwen-1.5B, and Phi-1.5, as described in Section \ref{experiment}. Every fine-tune shared the same hyperparameters. While Llama and Deepseek had almost the same performance, Phi needed a different LoRA Rank to make the percentage of training parameters closer to the others. For the latter, we increased the number of epochs as shown in Table \ref{tab:models_parameters} since it struggled in completing the task reaching the same performance as the others. LoRA $\alpha$ is 64 among all the models.

\begin{table*}[t]
    \centering
    \setlength{\tabcolsep}{4pt}
    \begin{tabular}{lcccc}
    \hline
        \textbf{Model} & \textbf{N param.} & \textbf{\% param. trained} & \textbf{LoRA $r$} & \textbf{Epochs} \\ \hline
        Llama-3.2-1B & 1.24B & 6.80 \% & 128 & 3 \\
        DeepSeek-R1-Distill-Qwen-1.5B & 1.78B &  8.73 \% & 128 & 3 \\
        phi-1\_5 & 1.42B & 5.43 \% & 256 & 6 \\ \hline
    \end{tabular}
    \caption{ Overview of the models models parameters used in our experiments, including their number of parameters, rank, and number of training epochs.
Hyperparameters such as target modules, $\alpha$ value, dropout rate, learning rate are held constant across all configurations and are detailed Table \ref{tab:appendix_hyperparameters} in Appendix \ref{sec:appendix}}
\label{tab:models_parameters}
\end{table*}

 \subsection{Ablation studies}
We conducted an ablation study to evaluate the model's performance without fine-tuning. The results showed that without fine-tuning, all models performed poorly, with accuracy ranging from 20\% to 30\%. This contrast highlights the essential role of fine-tuning in enabling the model to perform the task, specifically in processing implicit representations, achieving an accuracy of approximately 90\% when implicit data is shown during the fine-tuning process.


\section{Results and Discussion} \label{sec:discussion}

\begin{table*}[t]
    \centering
     \setlength{\tabcolsep}{4pt} 
    \begin{tabular}{lccccc}
    \hline
        \textbf{Mode} & \textbf{Acc.} & \textbf{Bal. Acc. } & \textbf{Precision} & \textbf{Recall} & \textbf{F1} \\ \hline
        Train and test explicit & 0.888 & 0.922 & 0.889 & 0.922 & 0.903 \\ 
        Train and test implicit & 0.911 & 0.914 & 0.890 & 0.914 & 0.900 \\ 
        Train explicit implicit, test explicit & 0.892 & 0.928 & 0.892 & 0.928 & 0.907 \\ 
        Train explicit implicit, test implicit & \textbf{0.933} & \textbf{0.947} & \textbf{0.915} & \textbf{0.947} & \textbf{0.930} \\ 
        Train explicit, test  implicit & \textbf{0.716} & \textbf{0.636} & \textbf{0.862} & \textbf{0.636} & \textbf{0.686} \\ \hline
    \end{tabular}
    \caption{Results on Llama 3.2-1B}
    \label{llama1B}
\end{table*}

\begin{table*}[ht!]
    \centering
    \setlength{\tabcolsep}{4pt} 
    \begin{tabular}{lccccc}
    \hline
        \textbf{Mode} & \textbf{Acc.} & \textbf{Bal. Acc.} & \textbf{Precision} & \textbf{Recall} & \textbf{F1} \\ \hline
        Train and test explicit & 0.883 & 0.923 & 0.882 & 0.923 & 0.900 \\ 
        Train and test implicit & 0.896 & 0.864 & 0.884 & 0.864 & 0.873 \\
        Train explicit implicit, test explicit & 0.900 & 0.939 & 0.897 & 0.939 & 0.915 \\ 
        Train explicit implicit, test implicit & \textbf{0.907} & \textbf{0.894 }& \textbf{0.891} & \textbf{0.894} & \textbf{0.891} \\
        Train explicit, test implicit & \textbf{0.671} & \textbf{0.588} & \textbf{0.732} & \textbf{0.588} & \textbf{0.598} \\ \hline
    \end{tabular}
    \caption{Results on DeepSeek R1 Distill Qwen-1.5B}
    \label{DeepSeek-1.5B}
\end{table*}

\begin{table*}[ht!]
    \centering
    \setlength{\tabcolsep}{4pt}
    \begin{tabular}{lccccc}
    \hline
        \textbf{Mode} & \textbf{Acc.} & \textbf{Bal. Acc.} & \textbf{Precision} & \textbf{Recall} & \textbf{F1} \\ \hline
        Train and test explicit & 0.889 & 0.906 & 0.899 & 0.906 & 0.902 \\
        Train and test implicit & 0.911 & 0.884 & 0.921 & 0.884 & 0.900 \\
        Train explicit implicit, test explicit & 0.896 & 0.925 & 0.897 & 0.925 & 0.910 \\
        Train explicit implicit, test implicit &\textbf{ 0.925} &\textbf{ 0.921} & \textbf{0.921} & \textbf{0.921} & \textbf{0.921} \\
        Train explicit, test implicit & \textbf{0.581} & \textbf{0.382} & \textbf{0.903} & \textbf{0.382} & \textbf{0.415 }\\ \hline
    \end{tabular}
    \caption{Results on Phi 1\_5B}
    \label{phi1-5}
\end{table*}

This work aimed to answer two main research questions. Regarding RQ1: \textit{How do implicit and explicit verbalizations affect LLM performance in information extraction tasks?} we evaluated how well a language model (GPT-4o-mini) extracted target information from both implicit and explicit textual data. Specifically, we measured the semantic distance between the model's predictions and the ground truth using Sentence-BERT. This yielded two sets of distance scores: one for explicit inputs and one for implicit inputs. A statistical comparison (Wilcoxon signed-rank test) between the two distributions revealed significantly higher distances for implicit descriptions, indicating that the model struggled more when information was conveyed indirectly. Supporting this, the analysis in Section~\ref{sec:preliminary_results} highlights two patterns: (1) a higher rate of failure cases (14.6\% ‘NaN’ values) for implicit texts compared to explicit ones (1.3\%); and (2) a greater frequency of low semantic similarity scores (BLEURT distance below 0.6) in the implicit condition. These results suggest areas where the model’s ability to handle indirect language remains limited. 
These findings indicate areas for improvement in IE tasks, which are explored further in RQ2: \textit{How does exposure to implicit data during fine-tuning affect an LLM’s ability to generalize to implicit reasoning tasks?}

Results shown in Tables [\ref{llama1B}, \ref{DeepSeek-1.5B}, \ref{phi1-5}] demonstrate that models trained on both explicit and implicit data consistently outperform those whose training rely only on explicit data when tested on implicit reasoning tasks. For instance, the Llama 3.2-1B model, fine-tuned on both types of data and tested on implicit tasks, achieved an accuracy of 93.3\%, a balanced accuracy of 94.7\%, and an F1 score of 93.0\%. These results show that exposure to both explicit and implicit verbalization increases the model's ability to generalize effectively across reasoning types.

Contrastingly, when models were trained on explicit data only, their performance on implicit data was significantly worse. For example, on Llama 3.2-1B, a model trained only on explicit data and tested on implicit data achieved an accuracy of only 71.6\%, and other performance metrics such as recall and F1 also suffered. Similarly, on DeepSeek R1 Distill Qwen-1.5B and Phi 1\_5B, models trained on explicit data showed similar difficulties, with accuracy dropping to 67.1\% and 58.1\%, respectively, when tested on implicit data.

In summary, the results demonstrate the effects of fine-tuning on LLMs for implicit reasoning tasks. In particular, we observe that when models are tuned on both explicit and implicit data, they show high performance in inference for both cases. However, models trained exclusively on explicit data have significant difficulties when confronted with implicit tasks. These results are in line with the findings of RQ1.

It is indeed not surprising from the evidence in Tables [\ref{llama1B} ,\ref{DeepSeek-1.5B},\ref{phi1-5}] that if the model sees in the training phase and in the testing phase, the same data distributions (\textit{test and train on implicit, test and train on epxlicit}) it is able to perform well on the required task. This points towards the conclusion that this difficulty in implicit IE is due to poor exposure in the training phase of implicit texts, making a fine-tuning phase necessary when handling texts with implicit information. 

\section{Conclusion} 
\label{sec:conclusion}
The results suggest that LLMs' difficulty with implicit information may be primarily due to insufficient exposure to implicit patterns during training rather than an inherent limitation of the model architecture. This test was carried out on LLama3.2 1B, DeepSeekV1-DistilledQwen1B, and Phi1-5, popular models in the community used for classification and generation.
The successful improvement through fine-tuning proposes a practical path forward for adapting existing LLMs to better handle implicit information in specific domains, as in our biographical data case.

Future developments could explore how different types of implicit patterns influence the implicit information extraction task. 

\section*{Limitations} \label{sec:limitation}
The results of this work are limited to biographical data. While many other types of text could be analyzed, retrieving such datasets is not as straightforward as generating a synthetic one using a specific subset of Wikidata. An additional limitation is the synthetic generation of the dataset: it may not fully reflect the complexity of naturally occurring implicit information in human-generated language.


\bibliography{custom}

\appendix
\clearpage

\section{Appendix A}
\label{sec:appendix}

\begin{table}[ht]
    \centering
    \begin{tabular}{ll}
    \hline
        \textbf{target\_modules}  & "self\_attn.q\_proj", "self\_attn.k\_proj", "self\_attn.v\_proj", "self\_attn.o\_proj", \\ 
        ~ & "mlp.gate\_proj", "mlp.up\_proj", "mlp.down\_proj" \\ \hline
        \textbf{LoRA alpha} & 64 \\ \hline
        \textbf{LoRA dropout }& 0.15 \\ \hline
        \textbf{learning rate }& $3^{e-5}$ \\ \hline
    \end{tabular}
    \caption{Hyperparameters held constant across all model configurations.
For model-specific settings such as rank and number of training epochs, please refer to Table \ref{tab:models_parameters} in the main text.}
    \label{tab:appendix_hyperparameters}
\end{table}

\end{document}